\documentclass{article}

\usepackage{PRIMEarxiv}

\usepackage{multirow}
\usepackage{amsmath}
\usepackage{float}
\usepackage{epsfig}

\title{Stochastic stem bucking using mixture density neural networks
}

\author{
  Simon Schmiedel \\
  Faculty of Science and Engineering \\
  Laval University \\
  Québec, Canada\\
  \texttt{simon.schmiedel.1@ulaval.ca} \\
}

\begin{document}
\maketitle

\begin{abstract}
Poor bucking decisions made by forest harvesters can have a negative effect on the products that are generated from the logs. Making the right bucking decisions is not an easy task because harvesters must rely on predictions of the stem profile for the part of the stems that is not yet measured. The goal of this project is to improve the bucking decisions made by forest harvesters with a stochastic bucking method. We developed a Long Short-Term Memory (LSTM) neural network that predicted the parameters of a Gaussian distribution conditioned on the known part of the stem, enabling the creation of multiple samples of stem profile predictions for the unknown part of the stem. The bucking decisions could then be optimized using a novel stochastic bucking algorithm which used all the stem profiles generated to choose the logs to generate from the stem. The stochastic bucking algorithm was compared to two benchmark models: A polynomial model that could not condition its predictions on more than one diameter measurement, and a deterministic LSTM neural network. All models were evaluated on stem profiles of four coniferous species prevalent in eastern Canada. In general, the best bucking decisions were taken by the stochastic LSTM models, demonstrating the usefulness of the method. The second-best results were mostly obtained by the deterministic LSTM model and the worst results by the polynomial model, corroborating the usefulness of conditioning the stem curve predictions on multiple measurements.
\end{abstract}

\keywords{bucking, harvester, LSTM, mixture density, neural network}

\section{Introduction}

\subsection{Research context}

Forest harvesters are a type of heavy machinery that fell trees and cut the stems into logs of predetermined dimensions. A harvesting head is mounted at the end of the boom which has a chainsaw, multiple sensors measuring the diameter and length of the stem and large knives to delimb the tree. Most modern harvesters are equipped with a bucking optimization software that selects the length of the logs that will be generated from the stem. The software is given a price matrix containing the dimensions of each log category and its respective weight (often referred to as price). Using dynamic programming, the software selects the combination of products that maximises the total sum of the weights of the products generated \cite{pnevmaticos1972} \cite{grondin1998}. When compared to manual selection of the products by the harvester operator, it was shown that using a bucking optimization software improves the value of the logs generated from forest stands \cite{olsen1991} \cite{labelle2018}. There also exists bucking to order methods that try to match the production to specific product distributions during harvesting. 

To optimize stem bucking, it is necessary to predict the stem profile of the part of the stem which has not yet been measured by the harvesting head. This can be challenging as measurements along the stem are correlated and the predictive model used should condition its predictions using the previous measurements taken from that stem. These correlations also imply that any measurement error may be propagated to the unmeasured part of the stem \cite{westfall2016} \cite{arias2017} together with any bias of the predictive model. Another challenge of bucking optimization is the relationship between the predictive model and the optimization software. Because the stem curve predictions are used to optimize the decisions made by the bucking algorithm, the errors made by the predictive model should be evaluated according to their impact on the optimization results instead of the difference with the true diameter.

\subsection{Literature review}

The prediction of stem curves, diameter and volume was previously extensively done using taper equations. These equations, often specific to a single species, subspecies, or subgroup of trees, are a parametric way to predict the stem curve and estimate its volume using the diameter at breast height and other exogenous variables \cite{kozak2004}. Taper equations were used for a wide range of species, including \textit{Tsuga heterophylla} \cite{flewelling1993}, \textit{Pinus taeda} \cite{fang2000} \cite{trincado2006}, \textit{Pinus elliotti} \cite{fang2000}, \textit{Liriodendron tulipifera} \cite{jiang2005}, \textit{Pinus pinaster} \cite{rojo2005}, \textit{Abies nordmanniana} \cite{sakici2008}, \textit{Picea glauca} \cite{yang2009}, \textit{Larix gmelinii} \cite{liu2020} and \textit{Betula platyphylla} \cite{he2022}, only to name a few. Because taper equations were developed for a multitude of the commercial tree species across the world, we refer readers who are looking for an exhaustive review and history of taper equations to \cite{mctague2021} and \cite{salekin2021}. 

There are multiple challenges encountered when using taper equations to predict the stem curve for bucking optimization. First, many traditional taper equations were created for forest inventory purposes and do not condition their predictions on multiple measurements along the stem, reducing their performance when they are used for stem bucking as including additional diameter measurements was shown to improve the predictions \cite{kozak1998}. One strategy to alleviate this problem was to develop taper equations for smaller subsets of stems that have similar characteristics \cite{arias2015} \cite{yang2022}. While doing so can improve the performance of the predictions, it is still not equivalent to conditioning the predictions on the previous stem measurements as it is done by harvesters. A second challenge ignored by much of the parametric taper equation research is the estimation of the generalization performance of the methods studied. When quantifying the performance of the predictions, there can be differences between the performance estimates made from the stems that were used to fit the taper equation and the estimates made from stems not used to fit the equations \cite{deMiguel2012}. Furthermore, the effects of the prediction errors made by the parametric taper equations on the bucking decisions are typically ignored. 

A notable family of methods used for taper predictions is the spline functions, which allow to model a wider range of stem curves. Cubic smoothing splines \cite{lappi2006} \cite{kublin2013} and penalized mixed splines \cite{scolforo2018} have both been previously used to predict stem taper. Noteworthy for stem bucking is the work of \cite{nummi2004} and \cite{koskela2006} who used cubic smoothing splines that conditioned their predictions using the previously measured part of the stem. A downside of their method was that the errors made by the model could compound over time for specific stems, leading to a gradual increase in the uncertainty of the predictions along the unknown part of the stem. While their goal was to improve stem profile predictions for stem bucking, they also did not estimate the impact of their predictions on the bucking decisions themselves. 

While the methods discussed up to this point all rely on the measurements made by the harvester head to make their predictions, other strategies have been tested using different technologies not yet present on harvester, the most popular of these being laser. As it was shown that including additional diameter measurements along the stem can improve taper predictions \cite{kozak1998}, the usefulness of using lasers to acquire such measurement was investigated by \cite{cao2009} and \cite{cao2011}. Lasers have also been used to measure the full stem profile by \cite{liang2014} and \cite{li2023}, creating a point cloud of the stem. Furthermore, \cite{prendes2023} demonstrated that if those technologies became widely available on harvesters, they could improve the decisions made during the bucking optimization process. Finally, camera images can also be used to measure the stem profile of the stems during harvesting \cite{grondin2023}. While these technologies may improve stem bucking in the future, they are not yet available on modern harvesters, limiting their immediate usefulness.

In recent years, supervised learning methods became prevalent in predictive modeling. However, small neural networks were already used for taper prediction in 2005 \cite{diamantopoulou2005} and 2006 \cite{kaloudis2006}. Since then, multiple supervised learning techniques were adapted to predict stem taper. One method that had mixed success on this problem is random forest. While \cite{schikowski2018} observed that random forests performed better than old parametric models for \textit{Acacia decurrens}, \cite{nunes2016} found the opposite for multiple Brazilian species. Also, random forest had poor performance extrapolating when compared to parametric predictive methods for \textit{Pinus taeda} \cite{yang2020} and six species of hardwoods \cite{yang2023}. 

While they have been used for almost two decades in stem taper predictions, in recent years there was a renewed interest for neural networks to accomplish this task. Neural networks were observed to perform worse than parametric models for \textit{Pinus sylvestris} \cite{seki2023} and \textit{Tectona grandis} \cite{fernandez2022}, however they performed better for \textit{Acacia decurrens} \cite{schikowski2018}, \textit{Fagus orientalis} \cite{sakici2018}, \textit{Abies nordmanniana} \cite{sakici2018}, \textit{Pinus sylvestris} \cite{ozccelik2019}, \textit{Pinus taeda} \cite{bonete2019},  three \textit{Nothofagus} species \cite{sandoval2022}, seven species in Poland \cite{socha2020} and multiple Brazilian species \cite{nunes2016}. While having a greater capacity to model complex relationships than traditional parametric models, the neural networks that were developed for stem taper prediction are not ideal for stem bucking. Like their parametric counterparts, they often do not condition their prediction on the measurements made previously on the stems as it is done during harvesting. They also do not evaluate their performance on the bucking itself but rather on the prediction errors.

Outside of the field of taper prediction, there exists a variety of neural network architectures that exploit the correlation between sequences of dependant variables. A method still relevant to this day is the Long Short-Term Memory network (LSTM) \cite{hochreiter1997} which improved the gradient propagation of previous recurrent networks allowing to handle longer sequences of data. This architecture was later modified by \cite{cho2014}, reducing the number of parameters that need to be trained. Recently, the Transformer architecture \cite{vaswani2017} has been used to make predictions successfully using very complex sequences of data by leveraging an attention mechanism that can give a greater importance to specific parts of the sequence. 

Another element which may be beneficial to neural network architectures used for stem bucking is the category of predictions made by the model. Instead of predicting single values for the stem taper as the previous research on taper prediction did, it is possible to predict the parameters for one or multiple statistical distributions using mixture-density neural networks \cite{bishop1994}. Using the negative log-likelihood of the distribution as the loss function, the predictions can be evaluated on the predicted probability density at the observed value. A bucking optimization algorithm that leverages the uncertainty of the predictions made might perform better that one that assumes the prediction are exact, reducing the propagation of prediction errors through the stem. 

\subsection{Research objectives}

The current state of the literature on stem profile predictions for bucking optimization highlights the need for a method that would condition its predictions on the previous parts of the stem while leveraging the uncertainty of the predictions to improve the bucking decisions. The usefulness of this method should be evaluated with its impact on the bucking decisions taken instead of with the prediction errors made. 

The goal of this project is to improve the bucking done by forest harvesters via a novel stochastic stem bucking method. Specifically, we aimed to:

\begin{itemize}
  \item Develop a predictive neural network that conditions its predictions on the previous measurements along the stem and predicts a statistical distribution of the unmeasured stem profile.
  \item Create a stochastic stem bucking algorithm to leverage the randomness of stochastic stem profile predictions. 
  \item Estimate the impact of stochastic stem predictions on bucking decisions.
  \item Distinguish the proportion of the bucking gains attributable to the use of the neural network to the gains attributable to the stochastic bucking decisions.
  \item Evaluate the sensitivity of the algorithm to the prices and dimensions of the products that can be created during the stem bucking.
\end{itemize}

\section{Materials and methods}

\subsection{Stem taper data}

The 8146 merchantable stem profiles used in this study were collected by the Québec (Canada) provincial government between 1999 and 2009 as a part of the forest inventory program \cite{misc:tiges}. The four major species used for lumber production in eastern Canada were considered in this study, which are \textit{Picea mariana}, \textit{Picea glauca}, \textit{Abies balsamea} and \textit{Pinus banksiana}. A description of the number of stems and diameter at breast height (DBH) for each species is illustrated in Table \ref{tab:stems}. 

\begin{table} 
\centering
\caption{Description of the stems in the data set}
\label{tab:stems}
\begin{tabular}{lllll}
\hline
\multirow{2}{*}{Species} & \multirow{2}{*}{\begin{tabular}[c]{@{}l@{}}Number of \\ stems\end{tabular}} & \multicolumn{3}{c}{DBH (cm)} \\
& & min. & avg. & max. \\ \hline
\textit{Abies balsamea} & 3043  & 10.8    & 22.2    & 40.0 \\
\textit{Picea mariana} & 2797 & 9.3     & 19.3    & 42.3 \\
\textit{Picea glauca} & 1358 & 11.4    & 28.9    & 56.8 \\
\textit{Pinus banksiana} & 948 & 9.0     & 21.0    & 47.4 \\ \hline
\end{tabular}
\end{table}

\begin{figure}
\centering
  \includegraphics[scale=0.6, angle=90]{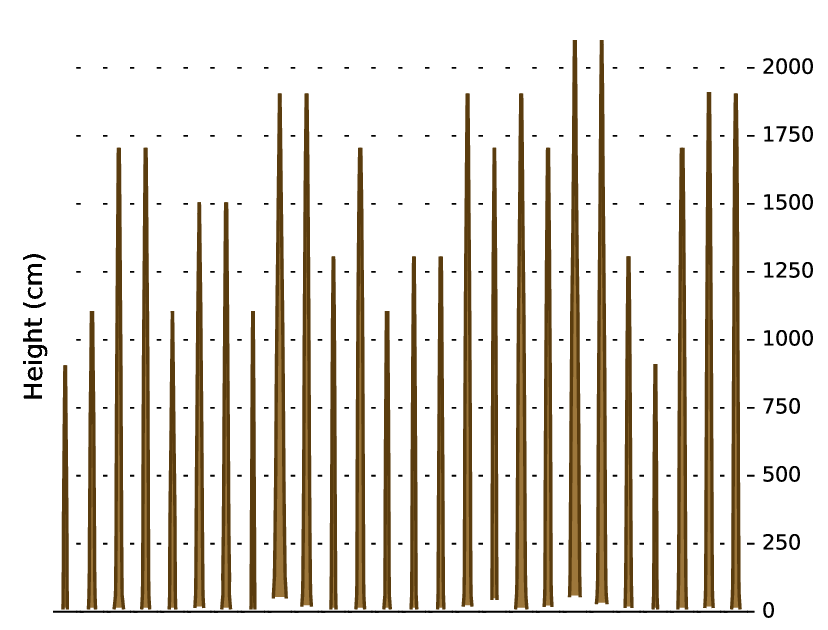}
  \caption{Sample of 26 stem profiles from the data set}
  \label{fig:stems}
\end{figure}

For every tree in the data set, diameter over bark measurements were manually taken at two-meter intervals along the stem with additional measurements close to the stump and at breast height. Contrary to measurements taken from harvesters which stop when the bucking ends, these stem profiles continue to the top of the tree in the part of the crown that is not merchantable. A sample of these stem profiles is illustrated in Figure \ref{fig:stems}. 

\subsection{Predictive taper model}

\subsubsection{Architecture}

The neural network architecture that was implemented to predict the stem profiles for stem bucking is the Long short-term memory network \cite{hochreiter1997}. Separate models were made for each of the four species. Because stems are relatively simple sequences of diameters, the LSTM is an ideal candidate as it has few parameters and good gradient propagation properties while still conditioning its predictions over the previous measurements along the stem. Given a sequence of stem diameters starting at the base of the tree, the network predicts the parameters of a normal distribution $\mathcal{N}(\mu, \sigma^2)$ for each subsequent diameter. 

The LSTM unit implemented in the network had a hidden size containing 10 features that led to an output vector also of size 10. An additional linear layer mapping the ten output features of the LSTM unit to a fully connected layer also of size 10 with a ReLU activation function and a second fully connected layer mapping to the size 2 output of the network $(\mu, \sigma^2)$ were added to further increase the representation capacity of the network.

\subsubsection{Loss function}

For the loss function of density-based models, \cite{bishop1994} recommended to use the negative likelihood. Because the likelihood of continuous distributions is based on the probability density function of that distribution, this implies that "good" predictions should predict a high density where the true values are and low densities elsewhere. The negative log-likelihood of the normal distribution $\mathcal{N}(\mu, \sigma^2)$ for $n$ observations can be simplified to what is displayed in Equation \ref{eq:simpli_res} (demonstrated in Appendix \ref{appendix1}), where $x_k$ is the true value of the $k^{th}$ observation.

\begin{equation} \label{eq:simpli_res}
\mathcal{L}_\mathcal{N} = \frac{1}{n} \sum_{k=1}^{n} (\ln(\sigma^2) + \frac{(\mu - x_k)^2}{\sigma^2})
\end{equation}

We can observe that the term on the left ($\ln{\sigma^2}$) penalizes the scenarios when the variance of the prediction is high, while the term on the right ($\frac{(\mu - x_k)^2}{\sigma^2}$) penalizes the scenarios when the squared error of the predictions is high. Increasing the predicted variance will increase the value of the term on the left while decreasing the value of the term on the right (and vice-versa). This loss function can be improved by adding a hyper-parameter $\lambda \in (0, 1)$ as is displayed in Equation \ref{eq:simpli_loss_final}. Adding this parameter allows to control the importance of the squared error and variance during back-propagation. This is the loss function that was used to train the stochastic neural networks.

\begin{equation} \label{eq:simpli_loss_final}
\mathcal{L}_\mathcal{N} = \frac{1}{n} \sum_{k=1}^{n} (\lambda\ln(\sigma^2) + (1-\lambda)\frac{(\mu - x_k)^2}{\sigma^2})
\end{equation}

\begin{figure} 
\centering
  \includegraphics[scale=0.5]{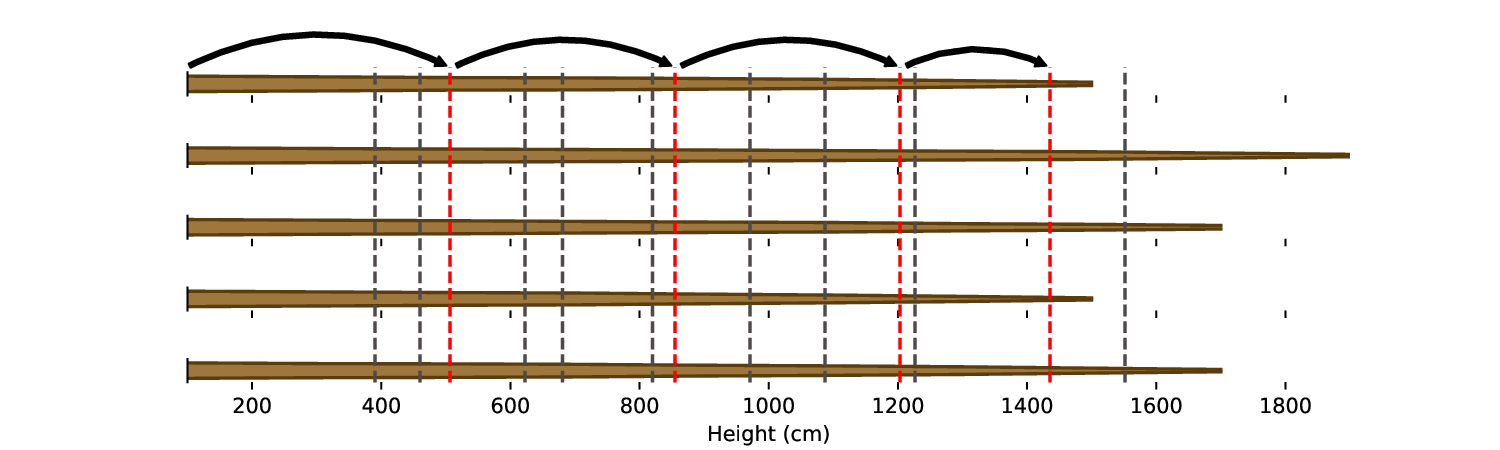}
  \caption{Stochastic bucking as a longest path problem using multiple predictions of the same stem profile. The dotted red lines indicate where the cuts maximizing the value of the products generated will be made, the arrows depict the corresponding longest path and the grey dotted lines are the cuts that were not chosen.}
  \label{fig:bucking_stem}
\end{figure}

\subsubsection{Benchmark models}

Two benchmark models were developed to compare the results obtained by the stochastic predictions made by the LSTM network. To assess the usefulness of stochastic predictions, a deterministic LSTM model was created which predicted only the diameter value instead of the parameters of a normal distribution. Other than the size of its output, it had the same architecture as the stochastic LSTM model. Because LSTM networks have not been previously used for taper predictions, a third model was created to compare the performance of the two LSTM-based models to simpler models that do not condition its predictions on the previously measured part of the stem. Polynomial models were chosen as they are simple, they can represent continuous functions and there is a history of their use for taper prediction \cite{sakici2008} \cite{allen1991} \cite{hjelm2013}. The loss function used for the two benchmark models is the squared error loss, which is displayed in Equation \ref{eq:sq_loss}.

\begin{equation} \label{eq:sq_loss}
\mathcal{L}_2 = \frac{1}{n} \sum_{k=1}^{n} (\mu - x_k)^2
\end{equation}

\subsection{Stochastic bucking optimization} \label{section:bucking}

Predicting probability densities instead of single values for the unmeasured stem diameter enables random sampling from the predicted distribution. Instead of predicting one stem profile as it was done in previous research, a sample containing multiple possible future stem profiles is created which is used by a stochastic bucking algorithm to identify the bucking decisions that are optimal when applied to the entire sample. The degree of similarity between the predictions in the sample will be modulated by the $\lambda$ parameter introduced in Equation \ref{eq:simpli_loss_final}, as decreasing its value will increase the variance in the diameter predictions. 

The stochastic stem bucking algorithm that was developed is inspired by \cite{grondin1998} and \cite{todoroki1999}. It formulates the bucking problem as a longest path problem maximizing the value of the products generated from the stems in the sample of possible future stem profiles, as is illustrated in Figure \ref{fig:bucking_stem}. Adapting the existing bucking algorithms to handle multiple stems at the same time led to new challenges that are not present in the deterministic case. First, the predicted stem profiles in the sample may not be the same height and diameter along the stem. This can lead to scenarios where some products could be made in some stems but not in others. To account for this, the value of a cut that generates a non-feasible product for a stem in the sample is set to zero for that specific stem. To evaluate the value of the cut for the entire sample, the mean value of the new cut is computed across all stem predictions in the sample.

A second new challenge of stochastic bucking is the choice of the height at which the algorithm must stop bucking because the stems stop being merchantable. This choice is non-trivial in the stochastic setting because all stems in the sample have different heights where they stop being merchantable. To handle this problem, the algorithm continues bucking while it can generate a product for at least one stem in the sample. There is a minimum stem diameter and a maximum height to stop individual stem predictions, which were respectively set to 4 cm and 40 m. Since using a sample size of one stem is equivalent to using a deterministic bucking algorithm, the stochastic bucking algorithm can be viewed as a generalisation of the deterministic bucking algorithm.

\subsection{Experimental design}

First, the stem profiles were randomly split into a training (60\%), validation (20\%) and test (20\%) set for each species. Copies of each stem where the measured part of the stem ends at different height were added to each data set to emulate the scenario where prior bucking decisions were previously taken in the lower part of the stem. 

\subsubsection{Hyper-parameter tuning for stochastic bucking}
\label{section:hyper_nn}

\begin{figure} 
\centering
  \includegraphics[scale=0.6]{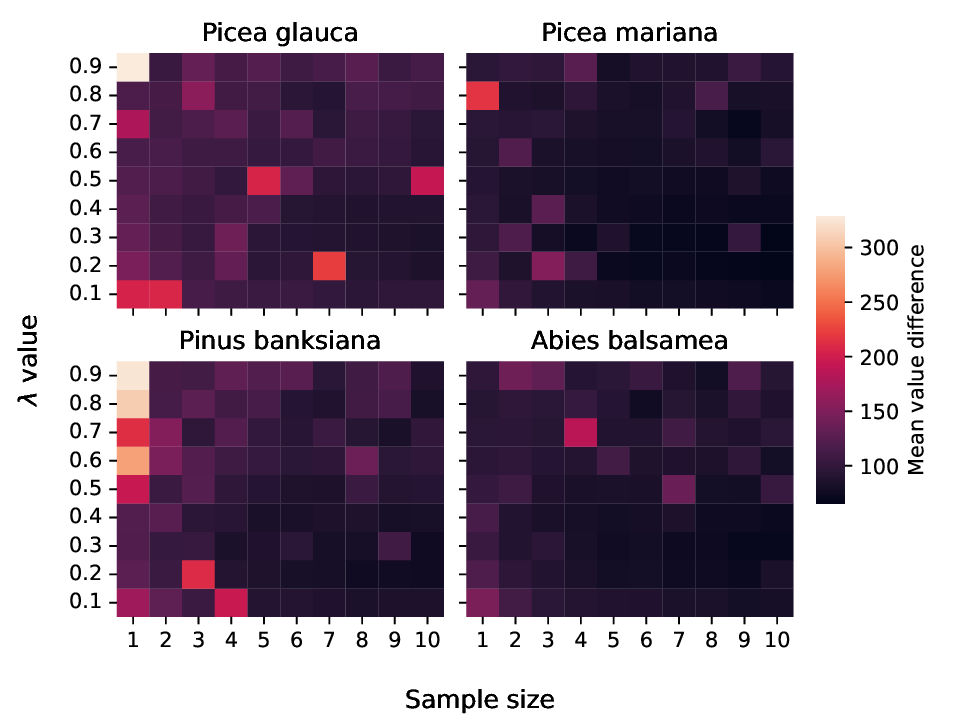}
  \caption{Mean difference in value of products generated on the validation data between the optimal bucking decisions and the decisions taken by stochastic bucking, according to the species, prediction sample size, and $\lambda$ value.}
  \label{fig:hyperparameter}
\end{figure}

To improve our understanding of the relationship between the hyper-parameter $\lambda$ of the loss function and the sample size hyper-parameter of stochastic bucking, multiple experiments were conducted. A challenge faced during hyper-parameter selection was the possible presence of relationships between the model hyper-parameters and other elements that may affect the performance of the bucking such as the dimensions and variety of the products that can be made, the values associated to them in the price matrix, the tree species, the tree size, etc. To reduce the number of experiments needed, five products with lengths of 251, 312, 373, 434 and 495 cm were included (corresponding to logs of 8, 10, 12, 14 and 16 feet with an extra margin), each having a minimum diameter of 9 cm and a maximum diameter of 100 cm. To allow the bucking algorithm to discard parts of the stem, a sixth product having a 30 cm length with no minimum and maximum diameter was added. To further reduce the number of experiments needed to study the effect of the hyper-parameters, a single price matrix was used. The price for each product was set to its length, except for the discarded product which had a price of 0. Using the lengths of the products as prices corresponds to maximizing the total length of the logs that are generated. This task may seem simple, however it was chosen as we hypothesized that hyper-parameters that performed well in this setting might perform well in other settings. 

A total of 360 experiments were conducted for hyper-parameter selection, each consisting of a specific combination of species, sample size, and $\lambda$ value. In each experiment a new neural network was trained. The bucking decisions made using the stochastic bucking algorithm were then compared to the bucking decisions made knowing the true stem profile. Since the best possible bucking decisions are done while knowing the true stem profile, the metric chosen to illustrate the results of the experiments was the difference between the value of the products generated using the true stem profile versus the value of the products generated using the stochastic predictions. The training details of these experiments are described in Section \ref{section:training} and the result are displayed in Figure \ref{fig:hyperparameter}. 

Across all four species, the differences between the total value of the products made knowing the true stem profile and the total value of product made using stochastic bucking ranged between $64.9$ and $328.9$, where a smaller difference is better. While they are not presented in the figure, 95\% confidence intervals over the mean difference were computed for each experiment, with a mean value of the confidence intervals across all experiments of $\pm 5.0$ and the largest value for any experiment being $\pm 21.0$. 

We can observe in Figure \ref{fig:hyperparameter} that for all four species, the hyper-parameter values where the value difference is the smallest seem to have a $\lambda$ between 0.2 and 0.4 and a sample size between 5 and 10. A lambda smaller than 0.5 means that more importance is accorded to reducing the squared error than to reducing the variance when training the neural networks. While this means that the variance of the predictions is not penalized as much, this increase in variance can be offset by the higher sample size in the stochastic bucking that led to better decisions overall. We can also observe that there is a visible variability present in the results which could be coming from the non-convexity encountered while training the neural networks. Finally, the average value difference was smaller for \textit{Picea mariana} and \textit{Abies balsamea} than for \textit{Picea glauca} and \textit{Pinus banksiana}. The hyper-parameter values selected for stochastic bucking for the remaining experiments were $\lambda=0.3$ and a sample size of 10 for all species.

\subsubsection{Training the deterministic LSTM model}

The deterministic LSTM model was evaluated with the same price matrix as in Section \ref{section:hyper_nn}. However, since the sample size and $\lambda$ value are not present in the deterministic model, no hyper-parameter tuning was done. For each species, a network was trained on the test data set and evaluated on the validation data set. The training details of these experiments are described in Section \ref{section:training}. The average difference in value with the best bucking decisions was then computed with 95\% confidence intervals. For \textit{Picea glauca}, this difference was $120.81 \pm 0.11$, for \textit{Picea mariana} it was $88.46  \pm 0.06$, for \textit{Pinus banksiana} it was $324.18 \pm 0.56$ and for \textit{Abies balsamea} it was $94.63 \pm 0.05$.

\subsubsection{Hyper-parameter tuning for the polynomial model}
\label{section:hyper_bench}

To create the polynomial model, the effect of the maximum order of the polynomial terms used in the regression was investigated. The hyper-parameter tuning of the polynomial model was conducted in a similar way as for the stochastic bucking model. For every combination of species and value for the maximum order of the polynomial terms, a new polynomial model was trained on the training data set. For every stem in the validation data set, the model predicted its stem profile using the first known diameter along the stem. These predictions were given to the bucking optimization algorithm and the resulting decisions were compared to the true optimal decisions. The products and price matrix used were the same as those used during the hyper-parameter tuning of the stochastic and deterministic LSTM bucking models. 

\begin{figure} 
\centering
  \includegraphics[scale=0.6]{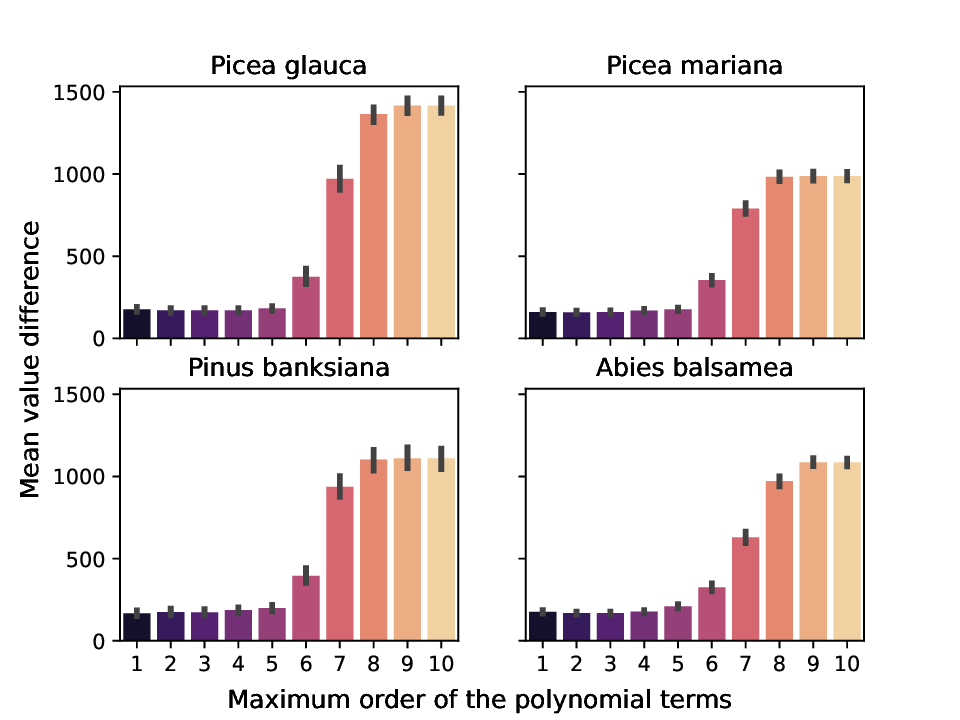}
  \caption{Mean difference in value of products generated on the validation data between the optimal bucking decisions and the decisions taken by the polynomial models, according to the species and maximum order of the polynomial terms. 95\% confidence intervals over the estimates of the means are displayed.}
  \label{fig:hyperparameter_benchmark}
\end{figure}

The differences in value of the products generated during stem bucking on the validation data between the true optimal decisions versus the decisions made using the prediction of the polynomial model are displayed in Figure \ref{fig:hyperparameter_benchmark}. For all four species, we can observe that the difference is smaller when the maximum order of the polynomial terms is between 1 and 5. However, including terms of order six or higher notably increased the differences. Because the models that had terms with maximum order between 1 and 5 seem mostly equivalent, we chose a maximum order of 1 for the final model. The exact difference observed with a maximum order of 1 were $176.5 \pm 14.7$ for Picea glauca, $160.6 \pm 10.2$ for Picea mariana, $166.7 \pm 17.7$ for Pinus banksiana and $175.6 \pm 9.8$ for Abies balsamea.

\subsubsection{Training details}
\label{section:training}

PyTorch 1.12.1. was used to implement the network and all algorithms were implemented in Python. The default initial weights were used as a starting point for the optimization. The learning rate was 0.001, and each model was trained for a total of 200 epochs using the Adam optimizer with a mini-batch size of 64.

\subsubsection{Bias and variance of the predictions}

The stochastic LSTM, deterministic LSTM and polynomial model were trained on the training data set using the hyper-parameters described in section \ref{section:hyper_nn} and \ref{section:hyper_bench} respectively. The models then predicted the stem profile of the stems in the test data set, and the bias and variance of the predictions were recorded (We are referring here to the true variance of the stem profile prediction, not the variance predicted in the loss function of the neural network).

\subsubsection{Effect of the minimum diameter of the products}

One factor that influences the bucking decisions taken is the minimum diameter of the products generated. While during the hyper-parameter tuning the minimum diameter was the same for every product in the price matrix, using different minimum diameters for the product dimensions could affect the performance of the bucking algorithms. To quantify the effect of the minimum diameter of the products on the bucking decisions, five scenarios were evaluated where the difference in the minimum diameter between smaller and bigger logs progressively increased. The same products lengths and values as in Section \ref{section:hyper_nn} were used, but with different minimum diameters for each products which are displayed in Table \ref{table:diam_scenarios}. For each model and scenario, the model was trained on the training data set. It then made predictions of the stem profiles of the test data set and made bucking decisions based on these predictions and on the scenario. The decisions taken by the models were then compared to the best possible decisions that are made when the true stem profile is known.

\begin{table} 
\centering
\label{table:diam_scenarios}
\begin{tabular}{lllllll}
\hline
\multicolumn{1}{c}{\multirow{2}{*}{\begin{tabular}[c]{@{}c@{}}Product\\ length \end{tabular}}} & \multicolumn{5}{c}{Minimum diameter scenario} & \multicolumn{1}{c}{} \\
\multicolumn{1}{c}{}                                                                          & 1          & 2      & 3      & 4      & 5     &                      \\ \hline
251                                                                                           & 9.00       & 9.00   & 9.00   & 9.00   & 9.00  &                      \\
312                                                                                           & 9.00       & 10.22  & 11.44  & 12.66  & 13.88 &                      \\
373                                                                                           & 9.00       & 11.44  & 13.88  & 16.32  & 18.76 &                      \\
434                                                                                           & 9.00       & 12.66  & 16.32  & 19.98  & 23.64 &                      \\
495                                                                                           & 9.00       & 13.88  & 18.76  & 23.64  & 28.52 &                      \\ \hline
\end{tabular}
\caption{Minimum diameter scenarios of the logs (values in cm)}
\end{table}

\subsubsection{Effect of the price of the products}

\begin{table} 
\label{table:price_scenarios}
\centering
\begin{tabular}{llllllllll}
\hline
\multicolumn{1}{c}{\multirow{2}{*}{\begin{tabular}[c]{@{}c@{}}Product\\ length \end{tabular}}} & \multicolumn{9}{c}{Price scenario} \\
\multicolumn{1}{c}{} & 1 & 2 & 3 & 4 & 5 & 6 & 7 & 8 & 9 \\ \hline
251 & 580.76 & 437.17 & 350.51 & 292.52 & 251.00 & 219.80 & 195.50 & 176.03 & 160.10 \\
312 & 545.76 & 443.97 & 382.54 & 341.44 & 312.00 & 289.88 & 272.66 & 258.86 & 247.56 \\
373 & 441.88 & 411.89 & 393.78 & 381.67 & 373.00 & 366.48 & 361.41 & 357.34 & 354.01 \\
434 & 269.12 & 340.91 & 384.24 & 413.24 & 434.00 & 449.60 & 461.75 & 471.48 & 479.45 \\
495 & 27.49 & 231.06 & 353.92 & 436.13 & 495.00 & 539.23 & 573.69 & 601.28 & 623.88 \\ \hline
\end{tabular}
\caption{Price of the products for the price scenarios.}

\end{table}

Another factor that has a direct effect on the bucking decisions taken is the price assigned to the products in the price matrix. To quantify the effect this has on the performance of the algorithms, nine price scenarios were created, each with different prices associated to the product discussed in Section \ref{section:hyper_nn}. The prices are not expressed in any specific currency. The specific prices of each products in these scenarios are displayed in Table \ref{table:price_scenarios}. While the scenario 5 uses the same prices as in Section \ref{section:hyper_nn}, scenarios 1-4 prioritize making smaller logs and scenarios 6-9 prioritize making longer logs. For each model and scenario, the model was trained on the training data set. It then made predictions of the stem profiles of the test data set and made bucking decisions based on these predictions and on the scenario. The decisions taken by the models were then compared to the best possible decisions that are made when the true stem profile is known.

\section{Results and discussion}

\subsection{Results}

\subsubsection{Bias and variance of the predictions}

\begin{figure} 
\centering
  \includegraphics[scale=0.6]{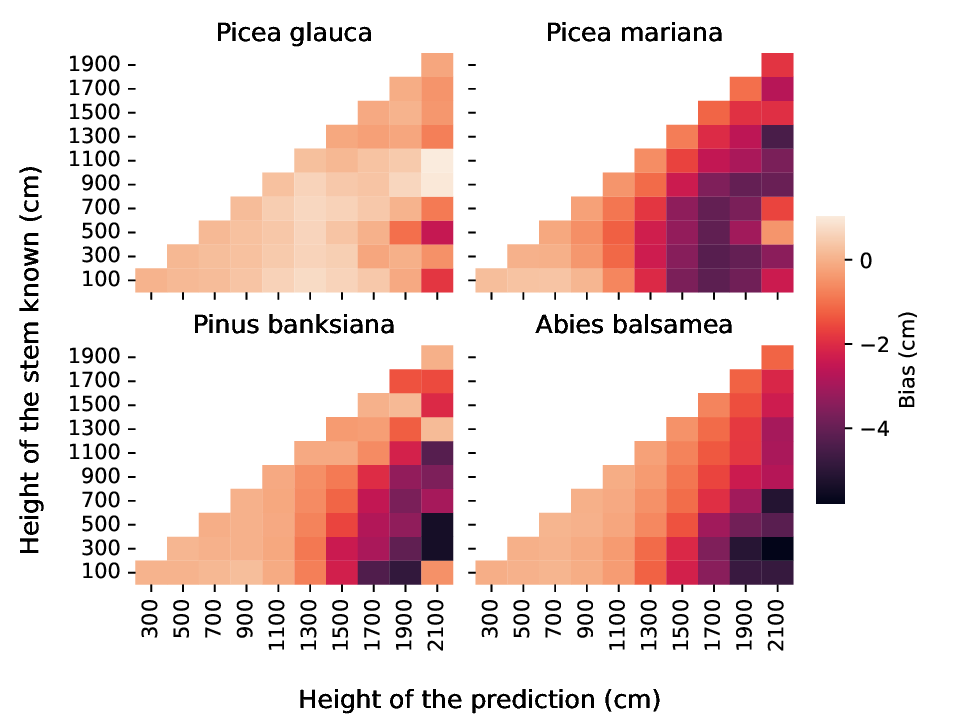}
  \caption{Bias of the predictions made by the stochastic LSTM models according to the height up to which the steam measurements were known and the height at which the predictions were made.}
  \label{fig:bias_NN}
\end{figure}

\begin{figure} 
\centering
  \includegraphics[scale=0.6]{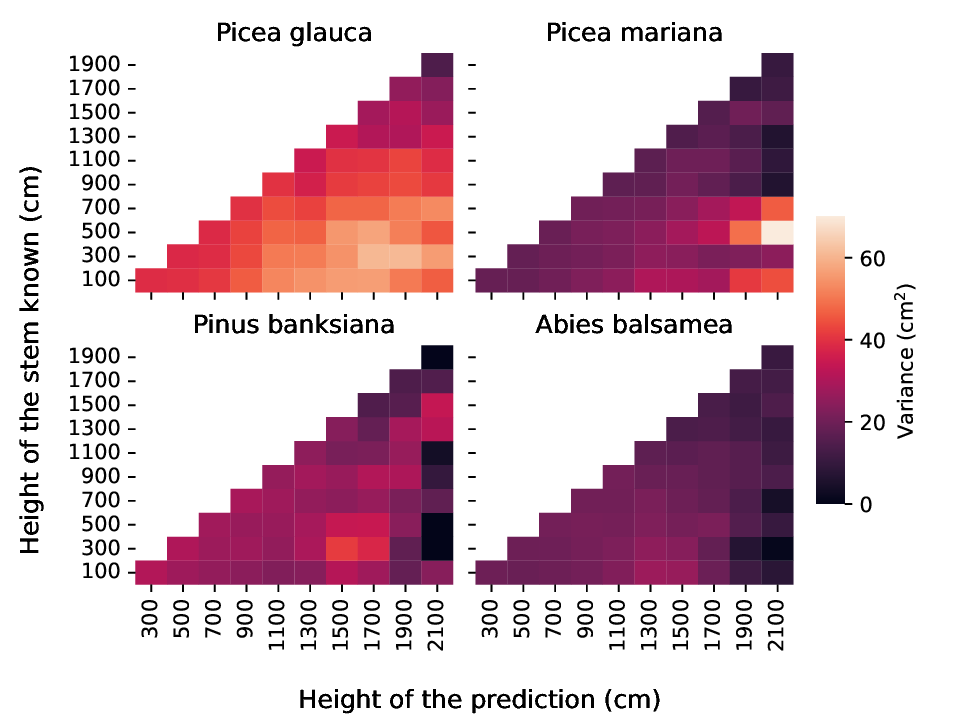}
  \caption{Variance of the predictions made by the stochastic LSTM models according to the height up to which the steam measurements were known and the height at which the predictions were made.}
  \label{fig:variance_NN}
\end{figure}

The bias and variance of the predictions made by the stochastic LSTM models on the test data set are displayed in Figure \ref{fig:bias_NN} and \ref{fig:variance_NN} respectively. While they are not displayed, 95\% confidence intervals for the bias and variance estimates were computed. The mean value of the confidence intervals for the bias estimates is $\pm 0.02$ cm and the maximum value is $\pm 0.43$ cm.  The mean value of the confidence intervals for the variance estimates is $\pm 0.12$ cm$^2$ and the maximum value is $\pm 1.30$ cm$^2$.

First, we can observe that for \textit{Picea mariana}, \textit{Pinus banksiana} and \textit{Abies balsamea}, the bias of the predictions became increasingly negative the further away the height of the prediction was from the last known measurement along the stem. Generally, the bias was closer to zero for predictions made in the lower part of the stems than for predictions made in the upper part of the stems. Furthermore, increasing the number of measurements known along the stem reduced the bias of the predictions. Finally, the bias of \textit{Picea glauca} was lower that the bias of the other species and did not seem to follow exactly the same trends. 

While trends for the variance of the predictions are not as apparent, we can observe that the variance for \textit{Picea glauca} is much higher than for the other species. This is noteworthy as the bias of \textit{Picea glauca} was generally lower than the bias of the other species and also behaved differently.

\begin{figure} 
\centering
  \includegraphics[scale=0.6]{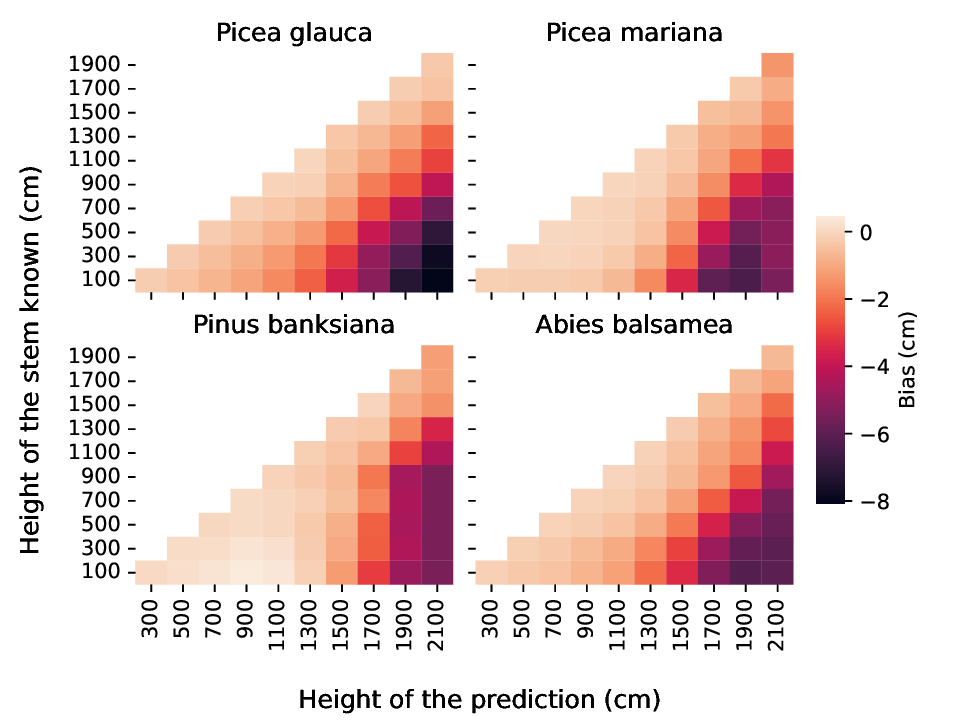}
  \caption{Bias of the predictions made by the deterministic LSTM models according to the height up to which the steam measurements were known and the height at which the predictions were made.}
  \label{fig:bias_detLSTM}
\end{figure}

\begin{figure} 
\centering
  \includegraphics[scale=0.6]{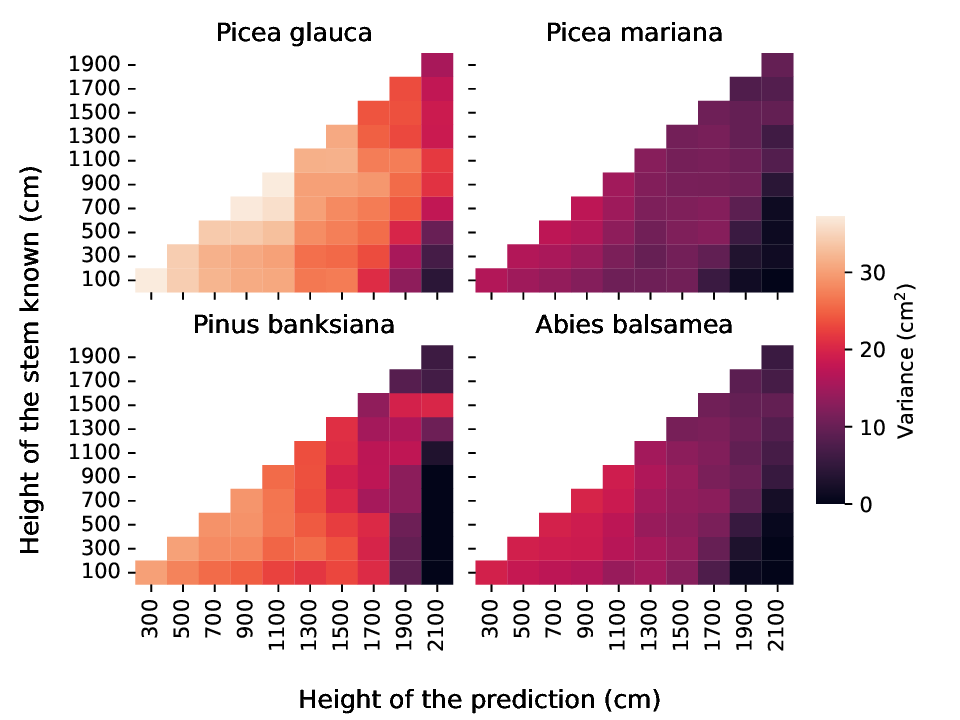}
  \caption{Variance of the predictions made by the deterministic LSTM models according to the height up to which the steam measurements were known and the height at which the predictions were made.}
  \label{fig:variance_detLSTM}
\end{figure}

The bias and variance of the predictions made by the deterministic LSTM models on the test data set are displayed in Figure \ref{fig:bias_detLSTM} and \ref{fig:variance_detLSTM} respectively. While they are not displayed, 95\% confidence intervals for the bias and variance estimates were computed. The mean value of the confidence intervals for the bias estimates is $\pm 0.01$ cm and the maximum value is also $\pm 0.01$ cm. The mean value of the confidence intervals for the variance estimates is $\pm 0.12$ cm$^2$ and the maximum value is $\pm 3.80$ cm$^2$. We can observe that for all four species, the bias of the predictions increased the further the predicted diameter was from the stump, while increasing the number of measurements that are known reduced the bias.

For the variance of the predictions, we can observe that \textit{Picea galuca} and \textit{Pinus banksiana} had  higher levels of variance, while \textit{Picea mariana} and \textit{Abies balsamea} had lower variance overall. For all four species, the variance of the predictions was higher for predictions made in the lower part of the stems, decreasing in the higher part of the stem. However, knowing more measurements along the stems increased the variance of the predictions.  

\begin{figure} 
\centering
  \includegraphics[scale=0.6]{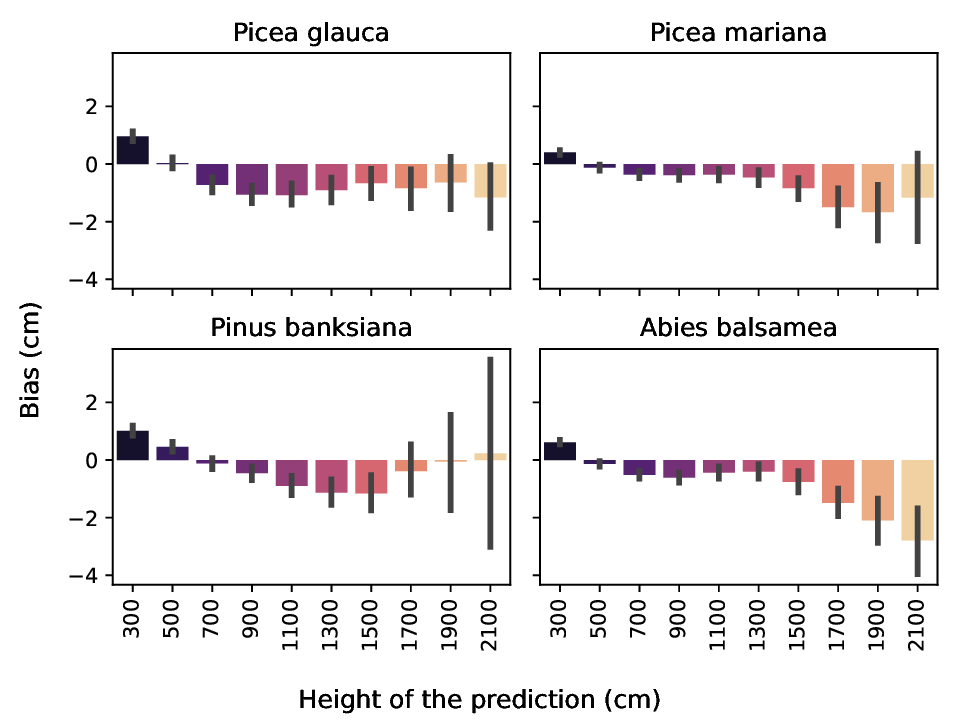}
  \caption{Bias of the predictions made by the polynomial models according to the height at which the predictions were made, with 95\% confidence intervals.}
  \label{fig:bias_bench}
\end{figure}

\begin{figure} 
\centering
  \includegraphics[scale=0.6]{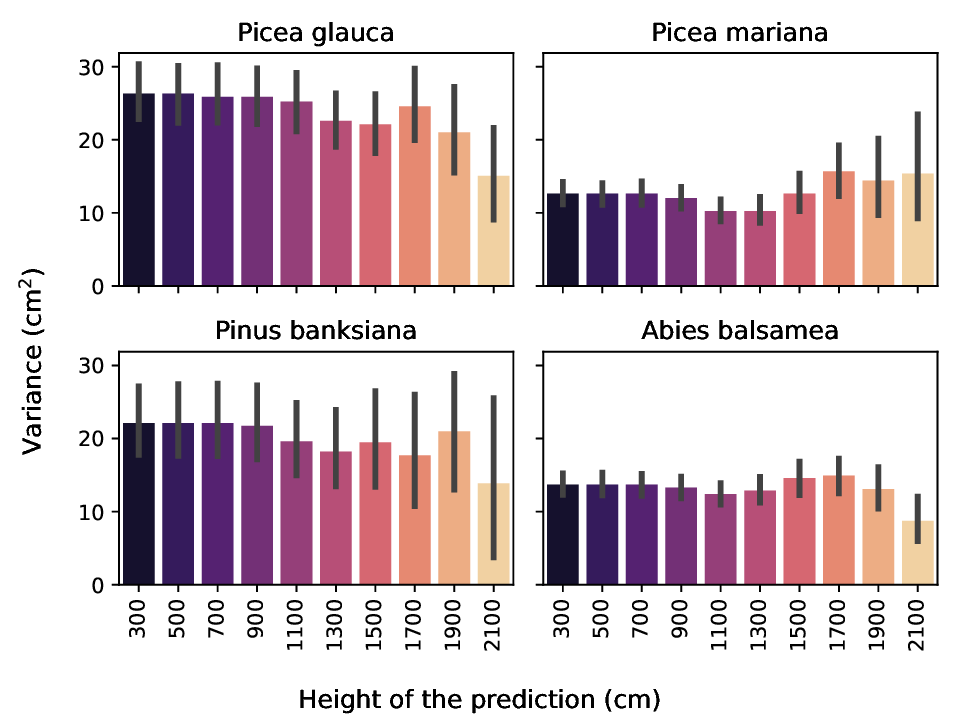}
  \caption{Variance of the predictions made by the polynomial models according to the height at which the predictions were made, with 95\% confidence intervals.}
  \label{fig:variance_bench}
\end{figure}

The bias and variance of the predictions made by the polynomial models on the test data set are displayed in Figure \ref{fig:bias_bench} and \ref{fig:variance_bench} respectively, with 95\% confidence intervals over the estimates. We can observe that generally the bias was positive when the height of the prediction is low and became negative as the height of the prediction increased. We can also observe that the magnitude of the confidence intervals increased together with the height of the prediction, especially at 1900 and 2100 cm.

For the variance of the predictions made by the polynomial models, the trends seem to vary amongst species. In general, the variance of the predictions at heights lower than 1100 cm was higher for \textit{Picea glauca} and \textit{Pinus banksiana} than for \textit{Picea mariana} and \textit{Abies balsamea}. At higher heights, the distinction is less evident because the amplitude of the confidence intervals increases.

\subsubsection{Effect of the minimum diameter of products}

\begin{table} 
\centering
\label{table:res_diam_min}
\resizebox{\textwidth}{!}{
\begin{tabular}{cllllll}
\hline
\multirow{2}{*}{Model} & \multicolumn{1}{c}{\multirow{2}{*}{Species}} & \multicolumn{5}{c}{Minimum diameter scenario} \\
 & \multicolumn{1}{c}{} & 1 & 2 & 3 & 4 & 5 \\ \hline
\multirow{4}{*}{Stochastic} & \textit{Picea glauca} & 87.74±0.08 & 74.36±0.08 & 67.57±0.08 & 59.22±0.07 & 53.55±0.07 \\
 & \textit{Picea mariana} & 67.72±0.04 & 52.49±0.04 & 41.27±0.04 & 34.97±0.04 & 28.08±0.03 \\
 & \textit{Pinus banksiana} & 105.28±0.18 & 80.21±0.16 & 67.87±0.15 & 63.44±0.14 & 52.61±0.13 \\
 & \textit{Abies balsamea} & 68.93±0.04 & 57.15±0.04 & 46.91±0.03 & 39.43±0.03 & 31.71±0.03 \\ \hline
\multirow{4}{*}{Deterministic} & \textit{Picea glauca} & 128.29±0.12 & 118.85±0.12 & 121.93±0.13 & 111.18±0.13 & 117.08±0.14 \\
 & \textit{Picea mariana} & 88.89±0.06 & 78.04±0.06 & 64.91±0.06 & 67.85±0.07 & 54.83±0.06 \\
 & \textit{Pinus banksiana} & 94.54±0.17
 & 85.29±0.18 & 81.0±0.19 & 74.19±0.19 & 67.74±0.18 \\
 & \textit{Abies balsamea} & 99.13±0.05 & 91.45±0.06 & 85.64±0.06 & 83.24±0.06 & 79.11±0.06 \\ \hline
\multirow{4}{*}{Polynomial} & \textit{Picea glauca} & 168.6±0.91 & 202.26±1.22 & 192.13±1.13 & 232.71±1.38 & 157.72±1.17 \\
 & \textit{Picea mariana} & 171.81±0.44 & 170.69±0.52 & 165.75±0.52 & 140.3±0.5 & 120.07±0.49 \\
 & \textit{Pinus banksiana} & 160.14±1.23 & 187.01±1.59 & 164.06±1.41 & 162.85±1.6 & 142.21±1.65 \\
 & \textit{Abies balsamea} & 165.74±0.39 & 185.78±0.47 & 174.44±0.44 & 168.91±0.52 & 159.25±0.53 \\ \hline
\end{tabular}}
\caption{Average value deviation from the optimal bucking solution according to the minimum diameter scenario, with 95\% confidence intervals.}
\end{table}

The results of the experiments on the effect of the minimum diameter of the products in the price matrix are displayed in Table \ref{table:res_diam_min}. We can observe that the bucking decisions taken by the polynomial models were in general much worse than the decisions taken by the stochastic LSTM models and deterministic LSTM models. The performance of the deterministic models was in general better than the polynomial models but worse than the stochastic models. We can also observe that increasing the differences in minimum diameter between smaller and bigger products improved the quality of the decisions made by the stochastic and deterministic models, however the same trend is not as visible for the polynomial models.

\subsubsection{Effect of the price of the products}

The results of the experiments on the effect of price of the products in the price matrix are displayed in Table \ref{table:results_cost}. We can observe that, in general, the deviation from the optimal solution decreases when more importance is given to creating shorter logs (scenarios 1-4) and increases when more importance is given to creating longer logs (scenarios 6-9). This trend is more apparent for the stochastic models than for the other models. We can also observe that the performance of the stochastic models is generally better that the performance of the polynomial and deterministic models. Furthermore, the deterministic models performed better than the polynomial model.

\begin{table} 
\label{table:results_cost}
\resizebox{\textwidth}{!}{
\begin{tabular}{cllllllllll}
\hline
\multirow{2}{*}{Model} & \multicolumn{1}{c}{\multirow{2}{*}{Species}} & \multicolumn{9}{c}{Price scenario} \\
 & \multicolumn{1}{c}{} & 1 & 2 & 3 & 4 & 5 & 6 & 7 & 8 & 9 \\ \hline
\multirow{4}{*}{Stochastic} & \textit{Picea glauca} & 16.95±0.08 & 29.74±0.08 & 52.9±0.08 & 77.32±0.08 & 87.24±0.08 & 104.53±0.1 & 104.42±0.11 & 106.24±0.12 & 108.66±0.12 \\
 & \textit{Picea mariana} & 12.93±0.05 & 25.27±0.05 & 42.32±0.04 & 57.05±0.04 & 68.0±0.04 & 77.05±0.05 & 80.29±0.06 & 81.69±0.06 & 81.97±0.06 \\
 & \textit{Pinus banksiana} & 0.43±0.02 & 12.58±0.08 & 32.28±0.08 & 55.27±0.11 & 75.5±0.14 & 81.95±0.17 & 83.28±0.19 & 81.19±0.19 & 78.4±0.19 \\
 & \textit{Abies balsamea} & 12.3±0.04 & 21.39±0.04 & 41.58±0.03 & 60.0±0.04 & 70.65±0.04 & 80.1±0.05 & 81.1±0.05 & 82.54±0.05 & 87.03±0.06 \\ \hline
\multirow{4}{*}{Deterministic} & \textit{Picea glauca} & 167.52±0.27 & 164.75±0.21 & 143.42±0.16 & 129.13±0.13 & 128.29±0.12 & 146.73±0.14 & 148.61±0.15 & 153.85±0.16 & 158.68±0.17 \\
 & \textit{Picea mariana} & 81.38±0.11 & 101.12±0.1 & 92.53±0.08 & 88.12±0.06 & 88.89±0.06 & 105.06±0.08 & 100.47±0.08 & 102.95±0.08 & 106.58±0.09 \\
 & \textit{Pinus banksiana} & 76.74±0.3 & 102.13±0.27 & 94.52±0.21 & 89.48±0.17 & 94.54±0.17 & 113.25±0.22 & 112.48±0.23 & 115.11±0.23 & 120.55±0.25 \\
 & \textit{Abies balsamea} & 115.66±0.11 & 122.64±0.09 & 109.9±0.07 & 99.88±0.06 & 99.13±0.05 & 116.33±0.07 & 115.65±0.07 & 119.28±0.07 & 124.09±0.08 \\ \hline
\multirow{4}{*}{Polynomial} & \textit{Picea glauca} & 123.84±2.03 & 168.21±1.68 & 150.24±1.17 & 142.53±0.89 & 168.6±0.91 & 232.36±1.2 & 222.97±1.28 & 232.41±1.36 & 225.03±1.42 \\
 & \textit{Picea mariana} & 131.94±0.96 & 189.36±0.81 & 168.09±0.58 & 154.93±0.44 & 171.81±0.44 & 217.38±0.58 & 207.09±0.62 & 212.79±0.65 & 221.48±0.69 \\
 & \textit{Pinus banksiana} & 165.06±2.98 & 207.76±2.37 & 180.27±1.63 & 156.5±1.2 & 160.14±1.23 & 211.13±1.58 & 213.05±1.73 & 222.98±1.81 & 223.72±1.94 \\
 & \textit{Abies balsamea} & 156.39±0.91 & 204.69±0.74 & 176.2±0.52 & 152.0±0.39 & 165.74±0.39 & 220.08±0.51 & 220.82±0.55 & 228.61±0.57 & 239.56±0.62 \\ \hline
\end{tabular}}
\caption{Average value deviation from the optimal bucking solution according to the price scenario, with 95\% confidence intervals}
\end{table}

\subsection{Discussion}

In the experiments conducted, the polynomial models were in general the worst models out of all model types. While the biases and variance of the predictions did not stand out much from the other models, the polynomial models had the disadvantage of not being able to condition their predictions on more than one measurement along the stems, which negatively affected the bucking decisions. Our results show that conditioning the predictions on more measurements is a crucial aspect of the stem curve prediction and should not be ignored by harvester manufacturers. These findings are in line with the results of \cite{koskela2006} who came to similar conclusions for their cubic smoothing spline model.

The deterministic LSTM model performed in general better in the experiments than the polynomial model but worse than the stochastic model. We believe that the first reason the results of the deterministic LSTM model were better than for the polynomial models was because the deterministic LSTM can conditions its predictions on multiple measurements along the stem. A second element that may have affected the results is the fact the LSTM is a neural network which is a more powerful model than the polynomial models, allowing it to better capture complex relationships in the stem profiles.

The stochastic LSTM models made better decisions than both the polynomial and deterministic LSTM models for most of the experiment conducted regardless of the species. In general, for the stochastic and deterministic models the bias increased, and the variance decreased the further away the predicted diameter was from the last known measurement along the stem. While for all models it is generally beneficial to have a low bias, the stochastic models had the advantage of alleviating the effect of a high variance by using multiple different predictions of the stem profile during bucking. The best value of the $\lambda$ hyper-parameter for the loss function reflected this as it favoured the reduction of the squared error over the reduction of the variance during training. We also observed that increasing the sample size never had a negative effect on the algorithm, however the gains progressively got smaller with each increase in sample size. The performance of stochastic bucking also highlights the importance of comparing the bucking decisions taken instead of the predictions made by the algorithms since increasing the variance of individual predictions can be beneficial in improving the bucking decisions while also increasing the prediction errors made.

During the experiments on the effect of the minimum diameter on the bucking decisions taken, we observed that when the minimum diameters diverged, the difference in value with the optimal solution decreased. We believe that this behaviour is caused by the increased constraints on the decisions that can be made during bucking. When the minimum diameter of a product increases, the number of possible bucking decisions can only decrease which could in turn reduce the difference between bucking decisions taken and the optimal solution.

During the experiments on the effect of the values of the logs in the price matrix, we observed a trend where the difference between the bucking decisions taken, and the optimal solution decreased when a higher price was assigned to the smaller logs and increased when a higher price was assigned to longer logs. We believe that this occurs because it is easier to make bucking decisions where smaller logs are favoured, and vice-versa. Because in a specific stem there are fewer long logs that can be made than smaller logs, the effect of predictions errors on the total value of the products generated from the stem is greater when the longer logs are favoured, and vice-versa.

There are some limitations to the results discussed here. First, the hyper-parameters of the stochastic and polynomial models were selected according to a specific price matrix. We cannot guarantee that we would have obtained the same values for these hyper-parameters if different products and prices were used in the price matrix. Changing the values of these hyper-parameters may affect our results and we believe that if these algorithms were implemented in another setting, new hyper-parameter values should be selected on the specific species and price matrix considered. Second, while we have studied the effect of the minimum diameter of the products and the effect of the value of these products separately, there may exist relationships between these two elements which we would have ignored here. While in this study we have considered the four main species of coniferous trees in eastern Canada, future research should also investigate the usefulness of stochastic bucking on different tree species prevalent in other parts of the world. Another promising area of research would be to integrate the decisions taken in the sawmills during the stochastic bucking optimization to directly maximize the value of the lumber generated by the stem instead of assigning a price to each log category.

\section{Conclusion}

Poor bucking decisions made by forest harvesters can have a negative economic impact on the viability of forest operations. Making the right bucking decisions is not an easy task because harvesters must rely on predictions of the stem profile for the part of the stems that is not yet measured. While previous research investigated stem tapers for a variety tree species, few efforts were made to develop models that conditioned their predictions on an unspecified number of diameter measurements along the stem and the resulting models were never evaluated by studying their impact on the bucking decisions taken. 

The goal of this project was to improve the bucking decisions made by forest harvesters with a novel stochastic bucking method. We developed a Long Short-Term Memory (LSTM) neural network which predicted the parameters of a Gaussian distribution, with which we can create a sample of stem profile predictions for the unknown part of the stem, conditioned on the known part of the stem. To do so we have adapted a loss function based on the Gaussian probability density function, allowing us to increase the importance given to reducing the squared error or the variance of the predictions during the training to the models. The bucking decisions could then be optimized using a novel stochastic bucking algorithm which made the bucking decisions over all the stems predictions in the sample.

The decisions made using stochastic bucking were compared to two benchmark models: A polynomial model that could not condition its predictions on more than one diameter measurement and a deterministic LSTM neural network. All models were evaluated on stem profiles of four coniferous species prevalent in eastern Canada (\textit{Picea galuca}, \textit{Picea mariana}, \textit{Pinus banksiana} and \textit{Abies balsamea}). In general, the best bucking decisions were taken by the stochastic models, demonstrating the potential of the method. The second-best results were obtained by the deterministic LSTM model and the worst results by the polynomial model, corroborating the necessity of conditioning the stem curve predictions on multiple measurements. 

Stochastic stem bucking showed great potential in improving the bucking decisions made by harvesters and future research should consider its usefulness for other commercial tree species.

\bibliographystyle{unsrt}  
\bibliography{paper}

\appendix
\section{Simplification of the negative log-likelihood of the normal distribution in a minimization context}
\label{appendix1}


\begin{equation*} \label{eq:simpli}
\begin{split}
\mathcal{L}_\mathcal{N} & = - \ln{\prod_{k=1}^{n} f_\mathcal{N}(x_k)}  = - \ln{\prod_{k=1}^{n} \frac{1}{\sqrt{2\pi\sigma^2}}}  
exp(-\frac{(\mu - x_k)^2}{2\sigma^2}) \\
\arg \min_{\mu, \sigma^2}\mathcal{L}_\mathcal{N} & = \arg \min_{\mu, \sigma^2} - \ln{\prod_{k=1}^{n} \frac{1}{\sqrt{2\pi\sigma^2}}}   
exp(-\frac{(\mu - x_k)^2}{2\sigma^2}) \\
& = \arg \min_{\mu, \sigma^2} - \sum_{k=1}^{n} \ln(\frac{1}{\sqrt{2\pi\sigma^2}} 
\exp(-\frac{(\mu - x_k)^2}{2\sigma^2})) \\
& = \arg \min_{\mu, \sigma^2} - \sum_{k=1}^{n} (\ln(\frac{1}{\sqrt{2\pi\sigma^2}}) + \ln(\exp(-\frac{(\mu - x_k)^2}{2\sigma^2})))\\
& = \arg \min_{\mu, \sigma^2} - \sum_{k=1}^{n} (\ln(\frac{1}{\sqrt{2\pi\sigma^2}}) - \frac{(\mu - x_k)^2}{2\sigma^2})\\
& = \arg \min_{\mu, \sigma^2} - \sum_{k=1}^{n} (\ln(1) - \ln(\sqrt{2\pi\sigma^2}) - \frac{(\mu - x_k)^2}{2\sigma^2})\\
& = \arg \min_{\mu, \sigma^2} - \sum_{k=1}^{n} ( - \ln(\sqrt{2\pi\sigma^2}) - \frac{(\mu - x_k)^2}{2\sigma^2})\\
& = \arg \min_{\mu, \sigma^2} - \sum_{k=1}^{n} ( - \frac{1}{2}\ln(2\pi\sigma^2) - \frac{(\mu - x_k)^2}{2\sigma^2})\\
& = \arg \min_{\mu, \sigma^2} - \sum_{k=1}^{n} ( - \frac{1}{2}\ln(2\pi) - \frac{1}{2}\ln(\sigma^2) - \frac{(\mu - x_k)^2}{2\sigma^2})\\
& = \arg \min_{\mu, \sigma^2} \sum_{k=1}^{n} ( \frac{1}{2}\ln(\sigma^2) + \frac{(\mu - x_k)^2}{2\sigma^2})\\
& = \arg \min_{\mu, \sigma^2} \sum_{k=1}^{n} (\ln(\sigma^2) + \frac{(\mu - x_k)^2}{\sigma^2})\\
& = \arg \min_{\mu, \sigma^2} \frac{1}{n} \sum_{k=1}^{n} (\ln(\sigma^2) + \frac{(\mu - x_k)^2}{\sigma^2})\\
\end{split}
\end{equation*}

\end{document}